\def\year{2021}\relax
\documentclass[letterpaper]{article} % DO NOT CHANGE THIS
\usepackage{aaai21}  % DO NOT CHANGE THIS
\usepackage{times}  % DO NOT CHANGE THIS
\usepackage{helvet} % DO NOT CHANGE THIS
\usepackage{courier}  % DO NOT CHANGE THIS
\usepackage[hyphens]{url}  % DO NOT CHANGE THIS
\usepackage{graphicx} % DO NOT CHANGE THIS
\usepackage{natbib}  % DO NOT CHANGE THIS AND DO NOT ADD ANY OPTIONS TO IT
\usepackage{caption} % DO NOT CHANGE THIS AND DO NOT ADD ANY OPTIONS TO IT
\usepackage{booktabs}
\usepackage{tabularx}
\usepackage{verbatim}
\usepackage{amsmath}
\usepackage{amssymb}
\title{An Approach to Improve Robustness of NLP Systems against ASR Errors}
\author{

    Written by AAAI Press Staff\textsuperscript{\rm 1}\thanks{With help from the AAAI Publications Committee.}\\
    AAAI Style Contributions by Pater Patel Schneider,
    Sunil Issar,  \\
    J. Scott Penberthy,
    George Ferguson,
    Hans Guesgen,
    Francisco Cruz,
    Marc Pujol-Gonzalez
    \\
}
\author{Tong Cui, Jinghui Xiao, Liangyou Li, Xin Jiang, Qun Liu}
\affiliations{
    Huawei Noah’s Ark Lab\\
    \{cuitong5, xiaojinghui4, liliangyou, Jiang.xin, qun.liu\}@huawei.com

}
\begin{document}

\maketitle

\begin{abstract}
Speech-enabled systems typically ﬁrst convert audio to text through an automatic speech recognition (ASR) model, and then feed the text to downstream natural language processing (NLP) modules. The errors of the ASR system can seriously downgrade the performance of the NLP modules. Therefore, it is essential to make them robust to the ASR errors. Previous work has shown it is effective to employ data augmentation methods to solve this problem by injecting ASR noise during the training process. In this paper, we utilize the prevalent pre-trained language model to generate training samples with ASR-plausible noise. Compare to the previous methods, our approach generates ASR noise that better ﬁts the real-world error distribution. Experimental results on spoken language translation(SLT) and spoken language understanding (SLU) show that our approach effectively improves the system robustness against the ASR errors and achieves state-of-the-art results on both tasks.
\end{abstract}

\section{Introduction}
In recent years, speech-enabled systems have become more and more widely used, particularly in the spoken dialog system and spoken language translation system. Voice assistants, like Amazon Alexa and  Apple Siri, are widely used in smartphones to obtain information and control devices. With simultaneous interpretation systems, people can hold a meeting with each other or watch live shows with automatically translated subtitles.  

Usually, a pipeline process is exploited to build the speech-enabled system. First, audio is converted into text by an automatic speech recognition (ASR) system. Then, the text is fed into downstream modules for different tasks. The errors of the ASR system in the first stage would propagate to the downstream modules and degrade their performance. \citet{belinkov2017synthetic} show that even a small perturbation in the ASR system could corrupt a machine translation (MT) system. Table \ref{table:asr_error_corrupt_mt} shows some practical examples of machine translation errors due to ASR noise.

\begin{table}[t]
	\centering
	\begin{tabular}{p{1cm}|p{5cm}}
		\toprule
		Speech    & Because I was having four heart attacks at the same time. \\
		\midrule
		ASR       & Because I was having. For heart attacks. At the same time,         \\
		\midrule
		Ref &    \begin{minipage}{.28\textwidth}
			\includegraphics[width=\linewidth, height=4mm]{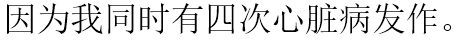}
		\end{minipage}      \\
		\midrule
		MT       & \begin{minipage}{.22\textwidth}
			\includegraphics[width=\linewidth, height=4mm]{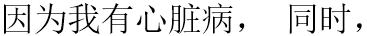}
		\end{minipage}   \\
		\bottomrule
	\end{tabular}
	\caption{An English-Chinese translation example with ASR errors. In this exmaple, ``having four'' is mis-recognized as ``haveing. For'', which causes the MT model produce erroneous translation.}
	\label{table:asr_error_corrupt_mt}
\end{table} 

A simple but effective way to deal with ASR errors is to train the downstream tasks with samples containing ASR noise.  It’s still a pipeline process without introducing additional inference time. \citet{di2019robust} use both clean manual transcriptions and noisy ASR outputs of audio paired with translation text to train MT models robust to ASR noise. However, such corpora are scarce and hard to obtain. To address this problem, data augmentation methods are used to generate training samples containing ASR-plausible errors.  Some heuristic rules are used in prior works, like homophones \cite{li2018improving}, similar pronunciation  \cite{tsvetkov2014augmenting} or confusion n-gram pairs harvested from aligned ASR-reference text pairs \cite{wang2020data}. However, these methods ignore the contextual information in sentences, making the generated samples sub-optimal. Moreover, a predefined threshold value is used for the ratio or amount of words to be replaced, which does not fit real-world ASR error distributions, such as the proportion of errors on different words.
\begin{figure*}[t]
	\centering
	\includegraphics[width=0.8\textwidth]{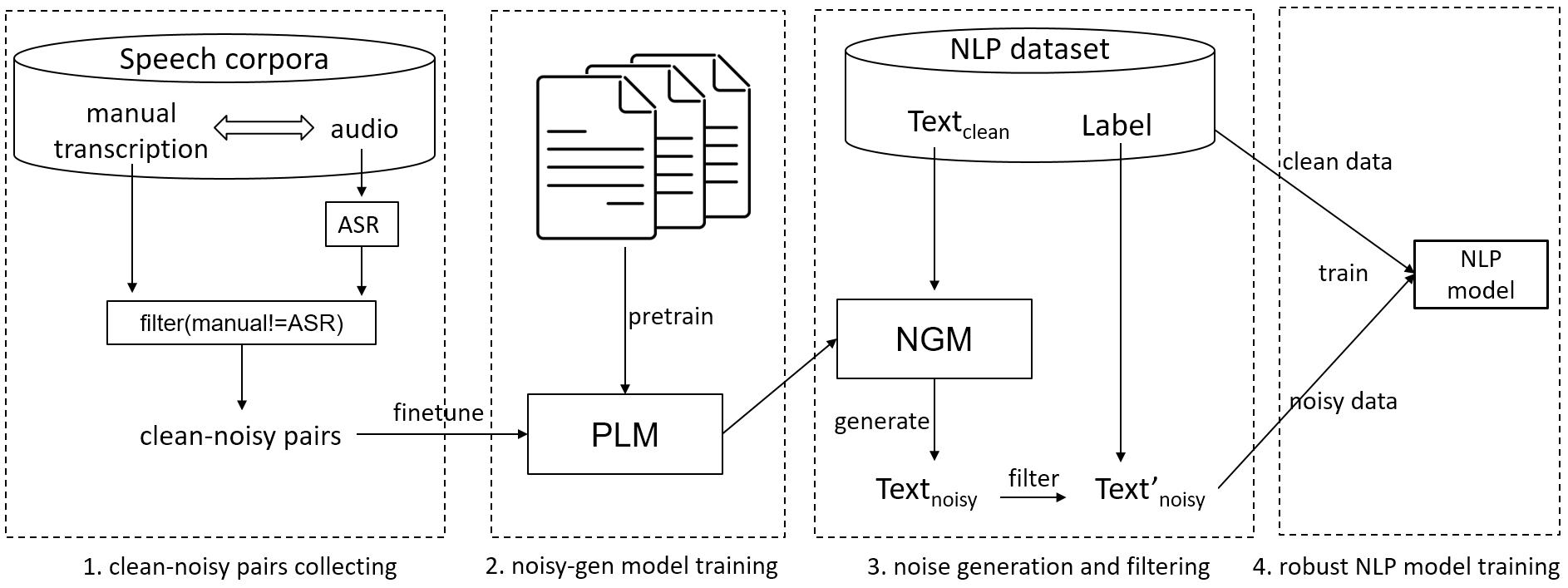}
	\caption{The illustration of our method. \texttt{PLM} refers to Pretrained Language Model. \texttt{NGM} refers to Noise Generation Model. Speech corpora must contain audio recordings and reference texts. ``NLP dataset'' refers to any dataset of supervised NLP tasks containing texts and their labels.  }
	\label{fig:asrnoisegen}
\end{figure*}

In this paper, we propose a novel data augmentation method leveraging a PLM to generate text containing ASR-plausible errors. Inspired by the success of PLMs on natural language generation tasks such as paraphrasing \cite{witteveen2019paraphrasing} and poetry generation \cite{liao2019gpt}, we fine-tune GPT-2 \cite{radford2019language} on a limited size of clean-noisy pairs to obtain a noise generator which is then used to add noise to input text of downstream tasks. The contributions of this paper are threefold.

\begin{itemize}

\item Firstly, we propose a novel data augmentation method using pre-trained language models to generate samples containing ASR-plausible errors.

\item Secondly, we conduct experiments on both SLT and SLU. To our best knowledge, this is the first work that applies to multiple downstream tasks. Experimental results show that our approach effectively improves system robustness against ASR noise and achieves state-of-the-art.

\item Thirdly, we contribute a large dataset containing ASR results of 3 popular public English speech corpus, recognized by a commercial ASR system. We will make the data publicly available soon.

\end{itemize}

\section{Background}
% Since our method is based on GPT-2, we will first introduce transformer and GPT-2 briefly before introducing our method.

Training language models has become a popular way of creating models suitable for transfer learning in the field of NLP \cite{mikolov2013distributed,peters2018deep}. While these models are initially trained in a self-supervised manner to predict the next word or the masked word in a sequence, they can be fine-tuned and used for a variety of downstream NLP tasks such as text classification, question answering, and text generation. 

More recently, large language models \cite{devlin2018bert,radford2018improving} using transformer  \cite{vaswani2017attention} architectures are achieving state-of-the-art results for many of these tasks while using less supervised data than previously needed. One of these large language models that has proven to be very good at text generation \cite{liao2019gpt,witteveen2019paraphrasing} is GPT-2  \cite{radford2019language}, which is a direct scale-up of GPT \cite{radford2018improving}, with more than 10X the parameters and trained on more than 10X the amount of data.

Here we will introduce transformer and GPT briefly before introducing our method.

\subsection{Transformer}
A standard Transformer layer
contains a Multi-Head Attention (MHA) layer and a Feed-Forward Network (FFN).
For the $t$-th Transformer layer, suppose the input to it is $X \in 
\mathbb{R}^{n\times d} $ where $n$ and $d$ are the sequence
length and hidden state size. Suppose there are $N_{H}$ attention heads in each layer, with head $h$
parameterized by $W^{Q}_{h}$, $W^{K}_{i}$, $W^{V}_{h}$, $W^{O}_{h} \in \mathbb{R}^{d\times d_{h}}$, and output of each head is computed as:

\begin{multline}
\mathrm{Attn}^{h}_{W^{Q}_{h}, W^{K}_{h}, W^{V}_{h}, W^{O}_{h}}(X)=\mathrm{Softmax}(\frac{QK^{T}}{\sqrt{d_{k}}})VW^{O}_{h} \\ =\mathrm{Softmax}(\frac{1}{\sqrt{d}}XW^{Q}_{h}W^{K\mathrm{T}}_{h}X^\mathrm{T})XW^{V}_{h}W^{O\mathrm{T}}_{h} 
\end{multline}

In multi-head attention, $N_{h}$ heads are computed in parallel to get the final output:

\begin{equation}
\mathrm{MHAttn}_{W^{Q}, W^{K}, W^{V},W^{O}}=\sum_{h=1}^{N_{h}}\mathrm{Attn}^{h}_{W^{Q}_{h}, W^{K}_{h}, W^{V}_{h}, W^{O}_{h}}(X).
\end{equation}

The FFN layer is parameterized by two matrices $W_{1} \in \mathbb{R}^{d \times d_{ff}} $ and $W_{2} \in \mathbb{R}^{d \times d_{ff}}$ where $d_{ff}$ is
the number of neurons in the intermediate layer of FFN. With a slight abuse of notation, we still use
$X \in R^{n\times d}$ to denote the input to FFN, the output is then computed as:

\begin{equation}
\mathrm{FFN}(X) = \mathrm{GeLU}(XW_{1} + b_{1})W_{2} + b_{2}
\end{equation}

\noindent where $b_{1}$, $b_{2}$ are the bias in the two linear layers.

\subsection{GPT}
Given an unsupervised corpus of tokens $U=\{u_{1},...,u_{n}\}$, a standard language modeling objective is to maximize the following likelihood:

\begin{equation}
L_{1}(U) = \log P(u_{i}|u_{i-k},...,u_{i-1};\theta) \label{1}
\end{equation}

\noindent  where $k$ is the size of the context window, and the conditional probability $P$ is modeled using a neural
network with parameters $\theta$. 

GPT uses a multi-layer transformer decoder for the language model, which is
a variant of the transformer. This model applies a multi-headed self-attention operation over the
input context tokens followed by position-wise feedforward layers to produce an output distribution over target tokens:

\begin{equation}
h_{0}=UW_{e} + W_{p} \label{2}
\end{equation}

\begin{equation}
h_{l}=\mathrm{transformer\_block}(h_{l-1}) \forall i \in [1,n] \label{3}
\end{equation}

\begin{equation}
P(u)=\mathrm{Softmax}(h_{n}W^{T}_{e}) \label{4}
\end{equation}

\noindent where $ U=(u_{k},...,u_{1})$ is the context vector of tokens, $n$ is the number of layers, $W_{e}$ is the token
embedding matrix, and $W_{p}$ is the position embedding matrix.

\section{Method}

Instead of using heuristics as previous data augmentation methods do, we make use of deep learning techniques, especially pre-trained language models, to generate samples with ASR noise. First, we collect clean-noisy pairs from speech corpora and the ASR results of them by pairing the erroneous ASR results with the golden transcriptions of the same audio. Then, we fine-tune GPT-2 on these pairs to get a noise generation model. Thirdly, we use the generation model to generate noisy input text of downstream tasks. Finally, we pair the generated noisy text with the ground truth label as augmented data and train a robust NLP model. Figure \ref{fig:asrnoisegen} illustrates the architecture of our method.

%Here we take machine translation as example, we feed a source sentence to the GPT model and generate several sentences containing ASR-pausible errors.

\subsection{Clean-noisy pairs collecting}

Firstly, we get the speech corpus, which includes both audio recordings of human speech and the corresponding transcriptions. Secondly, we put the audio into an ASR system to get the recognized results. Finally, we compare the ASR results with golden transcriptions in token-level and filter out exactly matched pairs.

\subsection{Noise generation model training}
The clean-noisy text pairs provide a supervised signal to train a generation model to inject ASR noise to input text. Recently, pre-trained language models show great capability and achieve significant improvements both in the discriminative task and the generative task of natural language processing. We take the GPT model \cite{radford2018improving} and fine-tune it on these clean-noisy pairs.  In detail, 
we concat clean text and noisy text with a \textit{[SEP]} token as the separator, and end it with a \textit{[EOS]} token as shown below:

\begin{figure}[h]
	\centering
	\includegraphics[width=0.5\textwidth]{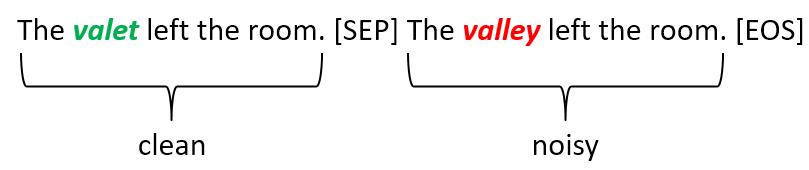}
	\caption{An example of training data for the noise generation model. } 
	\label{fig:ngm_train_example}
\end{figure}

The learning objective is to maximize the probability of observed sequence $ X=\{x_{1}, x_{2}, ...,x_{|x|}\} $:

\begin{equation}
P(X)=\sum\limits_{1\le i \le|X|}\log p(x_{i}|x_{1},...,x_{i-1}) \label{3}
\end{equation}

\noindent where $p(x_{i}|x_{1},...,x_{i-1})$ is the conditional probability of the token $x_{i}$ 
given all the historical tokens. The training is done in a small number of update steps to avoid overfitting the training data.  

\subsection{Noise generation and filtering}
After the PLM is fine-tuned, we use the model to augment the training data of downstream NLP systems.  
We feed clean text followed by a \textit{[SEP]} token into the model, and it generates noisy text token-by-token to get the noisy text. The process is ended when reaching the \textit{[EOS]} token.

During decoding, instead of beam-search, we apply Nucleus Sampling \cite{holtzman2019curious} strategy to improve the diversity of generation. At each step, only the most probable tokens with probabilities that add up to $p$ or higher are kept for generation, from which a specific token is sampled.  
%We take machine translation task as an example, as shown in Figure \ref{fig:asrnoisegen}. For the input source sentence: \textit{The priest tied the knot.}, the GPT model generates sentences similar in pronunciation but contains ASR-like noises, for example, \textit{The priest told the knot.} and \textit{The priest tied the night.}.

In practice, we find that too much generated noise may harm the performance of NLP systems and it is necessary to filter them out. We propose a scoring metric called Phone Edit Rate (PER) to evaluate the noisiness of the generated samples:

\begin{equation}
	\mathrm{PER}(O, G) = \frac{\mathrm{Levenshtein}(\mathrm{Phone}(O), \mathrm{Phone}(G))} {\mathrm{Len}(\mathrm{Phone}(O))} \label{3}
\end{equation}

\begin{table*}[h]
	\centering
	\begin{tabular}{c|c|c}
		\toprule
		$Text_{clean}$ & The priest tied the knot. & PER\\
		&\texttt{DH AH0 P R IY1 S T T AY1 D DH AH0 N AA1 T}& \\
		\midrule
		$Text_{noisy}$ & The priest told the knot.   & 0.13  \\   
		& \texttt{DH AH0 P R IY1 S T T OW1 L D DH AH0 N AA1 T}  &   \\
		\cline{2-3}
		& The priest down the knot    & 0.2  \\
		&  \texttt{DH AH0 P R IY1 S T D AW1 N DH AH0 N AA1 T}  &   \\ 
		\cline{2-3}
		 & The priest to you, you.    & 0.467   \\
		 & \texttt{DH AH0 P R IY1 S T T UW1 Y UW1 Y UW1}  &     \\ 
		\cline{2-3}
		& The priest tied the night. & 0.067   \\
		&\texttt{DH AH0 P R IY1 S T T AY1 D DH AH0 N AY1 T}  & \\ 
		\cline{2-3}
		& The priest tied the knot.Dot. & 0.2\\
		& \texttt{DH AH0 P R IY1 S T T AY1 D DH AH0 UNK} & \\ 
		\bottomrule
	\end{tabular}
	\caption{An example of noisy text generated by our method. In each cell, the first line is text, the second line is the phoneme sequcenes of the text transformed by cmu-dict. OOV words are transform to \texttt{UNK}.}
	\label{table:noise_gen_examples}
\end{table*}

\noindent where $O$ is the original text and $G$ is the generated text, $\mathrm{Phone}$\footnote{We use CMU Pronouncing Dictionary:\url{http://www.speech.cs.cmu.edu/cgi-bin/cmudict} for the transformation.} is a function that transforms a piece of text to its phonetic sequence. We calculate PER for each of the generated text, and filter out samples whose PER is larger than a given threshold. 
Table \ref{table:noise_gen_examples} shows an example of generated noisy texts and the PER value of each one.

\subsection{Robust NLP model training}
For each sentence in the clean training data, we generate $n_{aug}$ augmented sentences and pair them with the ground truth labels to get the noisy training data. Then we use both the clean and the noisy training data to train the NLP model such that the NLP model can tolerate real-world ASR noise. In this paper, we adopt two kinds of NLP models. One is machine translation, which is a classical sequence to sequence task. The other is language understanding, which we treat as a classification task. 

\section{Experimentas}

We evaluate our method on both SLT and SLU. We use a commercial ASR system throughout our experiments.

\subsection{Baselines}

\subsubsection{Rule-based confusion substitution (RS)} Following the work of \cite{li2018improving,tsvetkov2014augmenting}, we randomly replace words with similarly pronounced words to construct noisy samples. First, we transform a word to its phonetic sequence using cmu-dict, then we generate candidates to replace for each word whose edit distance of phonetic sequence is less than a threshold.

\subsubsection{Statistic-based confusion substitution (SS)} Following the work of \cite{wang2020data}, first we use scikit\footnote{\url{https://github.com/usnistgov/SCTK}} to align the pair of clean text and noisy text mentioned in 4.1. Then we extract confusion pairs and count their frequences. 

In both RS and SS, for a clean sentence, we randomly select a proportion of positions and replace the word/n-gram $w$ at each position with one of its candidates $\widetilde{w}$ by the following distribution:

\begin{equation}
	P(\widetilde{w}) = \frac{W(\widetilde{w})}{\sum\limits_{\widetilde{w}'\in V(w)}W(\widetilde{w}')}
\end{equation}

\noindent where $V(w)$ is the candidate set of $w$. $W$ is the weight of candidate. We count the term frequency from a large corpus as $W$ for RS and use the confusion frequency for SS. Table \ref{table:baseline_example} shows some examples of RS and SS.

\begin{table}[h]
	\centering
	%\begin{tabular}{p{1cm}|p{1cm}|p{1cm}}
	\begin{tabular}{c|c|c|l}
		\toprule
		&original& candidates & weight\\
		\midrule
		RS & good & \textit{could} & 534636\\
		&&  goode &70\\
		&& would &912456\\
		&&hood&3776\\
	    &&wood&26279\\
	    &&should& 885926\\
		\midrule
		SS & what is& uh what's & 1\\
		&& what's&15\\
		&& \texttt{EMPTY} & 11 \\
		&& and & 4 \\
		&& with& 2\\
		\bottomrule
	\end{tabular}
    \caption{Examples of RS and SS. \texttt{EMPTY} refers to empty string, which means the original n-gram is deleted. }
    \label{table:baseline_example}
\end{table}

\subsection{Implementation details}
\subsubsection{Clean-noisy pairs collecting}
We collect some popular speech corpora for our experiments. 
For English, we adopt Common Voice\footnote{https://commonvoice.mozilla.org/en/datasets},  tatoeba\_audio\footnote{https://tatoeba.org/eng/downloads} and  LJSpeech-1.1\footnote{https://keithito.com/LJ-Speech-Dataset/}. All of them contain the audio recordings of speech and their according transcriptions.

\begin{table}[h]
	\centering
	\begin{tabular}{cccc}
		\toprule
		Name               & \#Hours & \#Recordings & Language \\
		\midrule
		Common Voice       &1488 & 854,444      & En       \\
		tatoeba\_audio     &- & 300,076      & En       \\
		LJSpeech-1.1       &24 & 13,100       & En       \\
		\bottomrule
	\end{tabular}
	\caption{Statistic of collected speech corpora}
	\label{table:asrcorpus}
\end{table} 

We put the audio recordings of the above corpora into the ASR system and get recognized results.  When we compare ASR results with golden transcriptions, punctuation and casing errors are ignored. Finally, we obtain about 930k pairs of golden transcriptions and the text with ASR noise.

\begin{comment}

\begin{table*}[h]
	\centering
	\begin{tabular}{ccc}
		\toprule
		\multicolumn{1}{c}{error type} & \multicolumn{1}{c}{reference}             & asr                                        \\
		\midrule
		Omission                       & This convention is a natural.             & Convention is a natural.                   \\
		Phonetic                       & The quick fox jumped on the sleeping cat. & The quickbooks jumped on the sleeping cat. \\
		& So I would be the next to go.             & So I what we the next to go?               \\
		& She could not sit still.                  & She could not city still.                 \\
		\bottomrule
	\end{tabular}
	\caption{asr error examples}
	\label{table:asrerrorsample}
\end{table*}
\end{comment}

\subsubsection{Noise generation model training}
We use HuggingFace Transformers\footnote{\url{https://github.com/huggingface/transformers}} for model training and prediction. We fine-tune on GPT-2 small with 117M parameters released by OpenAI \cite{radford2019language}. The learning rate is set to 5e-5, and the batch size is set to 96. We train the model for 80,000 steps. Linear schedule is adopted for learning rate decay. We dump a model every 2000 steps and choose the model with the smallest perplexity on MSLT En-Zh dev set.

\subsubsection{Noise generation and filtering}
We set $p$ to 0.9 and softmax temprature to 1.0 for Nucleus Sampling. For PER caculation, we use cmu-dict to transform english sentences to phoneme sequences, out-of-vocabulary words to \textit{UNK}. Table \ref{table:noise_gen_examples} shows an example of generated text and corresponding phone sequences

\begin{table}[h]
	\centering
	\begin{tabular}{ccccc}
		\toprule  
		\multicolumn{1}{l}{} & \multicolumn{2}{c}{En-De}                     &  \multicolumn{2}{c}{En-Zh}                             \\
		\midrule  
		\multicolumn{1}{l}{} & dev                   & test                  & dev                   & test                       \\
		\midrule  
		CoVoST-2             & 15,531                 & 15,531                   & 15,531 & 15,531\\
		MSLT                 & 2,223                  & 2,128                     & 3,031       & 2,996        \\
		\bottomrule             
	\end{tabular}
	\caption{Statistics of the SLT test sets.}
	\label{table:slt_test_sets}
\end{table}

\begin{comment} 
\begin{table*}[h]
	\centering
	\begin{tabular}{cc}
		\toprule
		original     & generated   \\
		\midrule
		See, the house is full of smoke & See the house is full of smoke smoke \\
		& See the house is full of smoke.      \\
		& See the house is free of smoke.      \\
		& See the houses for smoke            \\
		\bottomrule
	\end{tabular}
	\caption{generated noise sample}
	\label{table:generatedsample}
\end{table*}
\end{comment}

.

\subsection{Experiments: Spoken Language Translation}
Following the work of \cite{di2019robust}, we first train a standard MT model with clean data. Then we augment the training data with noisy data generated by our method and fine-tune the MT model on the augmented data.
\subsubsection{Datasets}

We verify our approach on two SLT tasks: English-German and English-Chinese.

For En-De, we replicate the setup of \cite{vaswani2017attention}, based on WMT’14 training data with 4.5M sentence pairs. For En-Zh, we pre-process the WMT’17 training data following \cite{hassan2018achieving}\footnote{using preprocessing script at: \url{https://github.com/sanxing-chen/NMT2017-ZH-EN}} and obtain 19.4M sentence pairs.

\begin{table*}[h]
	\centering
	%\resizebox{\textwidth}{!}{
	\begin{tabular}{l|cc|ccc|cc|cc}
		\toprule
		& \multicolumn{5}{c}{En-Zh}  & \multicolumn{4}{c}{En-De}  \\ 
		\midrule
		& \multicolumn{2}{c}{CoVoST-v2}  & \multicolumn{3}{c}{MSLT}  & \multicolumn{2}{c}{CoVoST-v2}  & \multicolumn{2}{c}{MSLT}             \\ 
		\midrule
		& Manual & ASR & Manual & ASR & ASR-other & Manual & ASR  & Manual & ASR\\
		\midrule
		Clean & \textbf{50.5}  & 28.2  & \textbf{45.3} & 28.2 &34.1 & 34.0    & 17.2   & 26.2 & 13.9                             \\
		+RS    & 50.5     & 28.2     & 45.3    & 26.7    &31.6    & \textbf{34.1}    & 17.0    & 25.9   & 13.7             \\
		%RS   & -     & -     & -    & -    &48.8 &      & 33.6    & 17.0    & 25.9 & -             \\
		%RS   & -     & -     & -    & -    &48.8 &      & 33.6    & 17.0    & 25.9 & -             \\
		+SS   & 50.3   & 29.7  & 44.4 & 28.9  &35.8  & 34.0    & 18.0    & 25.6 & 14.3        \\ 
		\midrule 
		+Ours  & 50.2  & \textbf{30.8}  & 44.4 & \textbf{32.0} &\textbf{38.6} & 34.0 &\textbf{19.0}  & \textbf{26.6} & \textbf{20.2}            \\
		\bottomrule
	\end{tabular}%}
	\caption{BLEU scores for the results of MT models on manual transcriptions and ASR results of test sets. ``Clean'' refers to models trained on clean training data. ``+SS'', ``+RS'', ``+Ours'' refer to models trained on augmented data. ``ASR-other'' refers to the ASR results provided by MSLT dataset. }
	\label{table:slt_results}
\end{table*}

We test En-De and En-Zh on CoVoST-2 \cite{wang2020covost} and MSLT \cite{FedermannLewis2017}. 
Table \ref{table:slt_test_sets} shows statistics of the test sets. We use the same ASR system to recognize all of the test sets. In addition, MSLT En-Zh test set comes with its own ASR results, we also evaluate performance on that. For En-De, we measure case-sensitive tokenized BLEU \cite{papineni2002bleu}. For En-Zh,  we measure detokenized BLEU \cite{post2018call}\footnote{SacreBLEU hash: BLEU+case.mixed+lang.en-zh+numrefs.1+smooth.exp+test.wmt17+tok.zh+version.1.4.12} following \citet{hassan2018achieving}. 

\subsubsection{Settings}
All of the following experiments are carried out based on the Transformer base Model \cite{vaswani2017attention}.  

We train the En-De model following the settings of \cite{vaswani2017attention} and En-Zh following the settings of \cite{hassan2018achieving}. We train 100k steps for En-De and 300K steps for En-Zh. We average the 5 checkpoints performing best on the validation set to create the final model. During decoding, we use a beam size of 4 for En-De and 5 for En-Zh following \cite{ghazvininejad2019mask}. We train the MT models on 8 NVIDIA V100 GPUs.

In the data augmentation phase, we generate 5 noisy sentences for each source sentence and filter out sentences very noisy by setting PER threshold to 1.0, then we randomly select $n_{aug}=1$ sentence. For RS and SS, we set substitute proportion to 0.1 and generate 1 noisy sentence for each clean sentence as well.

Then, we continue training 100k steps for the En-De model and 200k steps for the En-Zh model on both clean and noisy training data.
%Our base models achive 34.4/23.7 bleu score on WMT17 En-Zh/Zh-En test set and 27.2 bleu score on WMT14 En-De testset, 

\subsubsection{Results}

The experimental results are shown in the Table \ref{table:slt_results}.

%\textbf{POINT1:conventional NMT models are not robust to ASR noise}

Firstly, the baseline model performs well on the test sets without ASR noise, but it suffers a great performance drop on the noisy test sets. The results are consistent with conclusions of prior work \cite{li2018improving,di2019robust} that the machine translation system is fragile to noise.

%\textbf{POINT2:our method can improve robustness to ASR noise, perform best }

\begin{figure}[h]
	\centering
	\includegraphics[width=0.5\textwidth]{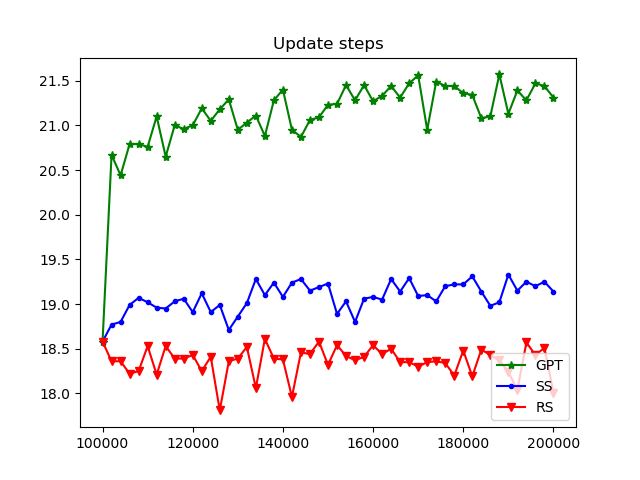}
	\caption{BLEU scores over update steps on the CoVoST-v2 En-De validation set. } % % }
	\label{fig:asrnoisegen}
\end{figure}

Secondly, our proposed method significantly outperforms both RS and SS on both En-De and En-Zh test sets with ASR noise. The performance of RS is equivalent or slightly worse than the baseline model, indicating that the ASR noise produced by RS does not fit the real-world ASR noise distribution. The performance of SS is better than the baseline model on all of the test sets because it harvests real ASR pairs, but the improvement is marginal due to the simple substitution strategy. Our method significantly improves the performance of MT models on ASR results, it improves BLEU score by 2.6/1.8 in CoVoST-v2 En-De/En-Zh, 3.8/6.3 in MSLT En-De/En-Zh. Figure \ref{fig:asrnoisegen} shows the evolution of BLEU scores during training.  At the same iteration, our results are consistently higher than both RS and SS. Moreover, our method achieves significant improvement (4.5 BLEU scores) on MSLT En-Zh with a different ASR system. It indicates that our method can produce some common ASR errors shared across different ASR systems, which can be used to improve model robustness to general ASR errors.

Lastly, in most cases, the models trained with augmented data will degrade performance slightly on clean input. The performance drop caused by our method is competitive to other data augmentation methods. We even get a better result on MSLT En-De clean set.  

\subsubsection{The effect of noisiness}
%\textbf{POINT3: the noise rate affect performance of model, analysis different PER threshold effect on model}
\begin{table}[h]
	\centering
	\begin{tabular}{c|c|cc|cc}
		\toprule
		&    & \multicolumn{2}{c}{CoVoST-v2}  & 	\multicolumn{2}{c}{MSLT}	 \\
		\midrule
		& $\alpha$ & Manual & ASR & Manual & ASR \\
		\midrule
		%& 0.1       & 34.1 &    17.5   &  25.9&     13.7 \\
		%+SS & 0.2       &34.0  &    17.4   &  25.2&     13.5 \\ 
		%& 0.3   & 34.0 & 17.0 &     25.8     &        13.6    \\
		%\midrule 	
		& 0.5       & 34.2 &    18.9   &  26.4&     19.7 \\
		+Ours& 1.0       &\textbf{34.4}  &    \textbf{19.0}   &  \textbf{26.6}&     \textbf{20.2} \\ 
		& $\inf$    & 34.1 & 18.9 &     25.9     &        18.6  \\ 
		\bottomrule	
	\end{tabular}
	\caption{Effect of different PER threshold $\alpha$ on BLEU scores of MSLT En-De test set.}
	\label{table:nmt_per_effect}
\end{table}	

In this section, we study the effect of the noisiness of augmented data on final performance. We use different PER threshold $\alpha$ to filter generated noisy texts. Larger $\alpha$ tends to involve noisier texts. The result is shown in Table \ref{table:nmt_per_effect}. We find that if the noisiness is too low, the model is less robust to ASR noise, if the noise is too much, the model performance downgrades on clean data and ASR results. The performance on MSLT test set is more sensitive to $\alpha$ than on CoVoST-v2.  

\subsubsection{A Case Study}

\begin{table}[h]
	\centering
	\begin{tabular}{p{1cm}|p{6cm}}
		\toprule
		Speech &  Yeah. Classic margherita pizza, besides the cauliflower.\\
		\midrule
		ASR & Yeah, class. Sick marguerite pizza. Besides the cauliflower, \\
		\midrule
		Ref &  \begin{minipage}{.3\textwidth}
			\includegraphics[width=\linewidth, height=4mm]{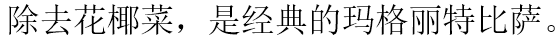}
		\end{minipage} \\
		\midrule
		\midrule
		Clean & \begin{minipage}{.35\textwidth}
			\includegraphics[width=\linewidth, height=8mm]{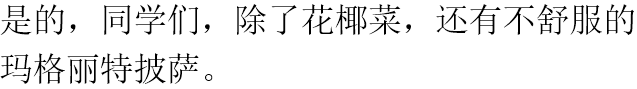}
		\end{minipage}\\
	\midrule
		+RS & \begin{minipage}{.35\textwidth}
			\includegraphics[width=\linewidth, height=8mm]{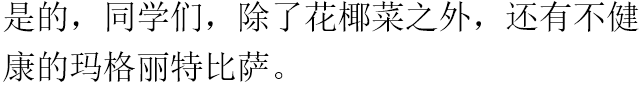}
		\end{minipage}\\
	\midrule
		+SS & \begin{minipage}{.35\textwidth}
			\includegraphics[width=\linewidth, height=8mm]{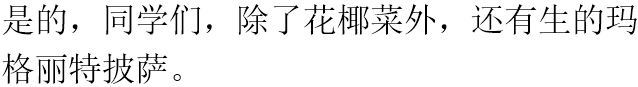}
		\end{minipage}\\
	\midrule
		+Ours & \begin{minipage}{.3\textwidth}
			\includegraphics[width=\linewidth, height=4mm]{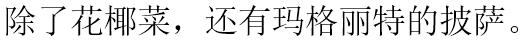}
		\end{minipage}\\
	\bottomrule
	\end{tabular}
	\caption{For the same erroneous ASR output, translations of the baseline En-Zh MT model and robustly trained MT models including 2 baseline data augmentation methods and our method.}
	\label{table:nmt_case_study}
\end{table}

In Table \ref{table:nmt_case_study}, we provide a realistic example to illustrate the
advantage of our robust MT system on erroneous ASR
output. In this case, the word "classic" is misrecognized as "class. Sick". 
The baseline MT model as well as other models trained with augmented data
can hardly avoid the translation of "class" and "sick" which are frequent characters with explicit word senses. In contrast,
our model avoids translating "class" and "sick" which will corrupt the translation completely and produces a more fluent and meaningful translation.
We consider that the robustness improvement is mainly
attributed to our proposed ASR-specific noise training.
\subsection{SLU}
\subsubsection{Datasets}
We use the following two SLU datasets. (1) Fluent Speech Commands (FSC) \cite{lugosch2019speech}: English speech commands related to personal assistant services.  (2) DSTC-2 \cite{henderson2014second}: human-computer dialogs in a restaurant domain collected using Amazon Mechanical Turk. We follow the same data preprocessing steps as in \cite{9053213,wang2020data} for DSTC-2. Statistics of the two datasets are summarized in Table \ref{table:stat_slu_datasets}. 

\begin{table}[h]
	\centering
	\begin{tabular}{lcccc}
		\toprule
		& \#Train & \#Dev  & \#Test & \#intents \\
		\midrule
		FSC   & 23,132 & 3,118 & 3,793 & 31 \\
		DSTC-2& 10,886 & 3,560 & 9,160 & 28 \\
		\bottomrule
	\end{tabular}
	\caption{Statistics of the SLU datasets.}
	\label{table:stat_slu_datasets}
\end{table}  
Each sample of FSC has exactly one intent, while DSTC-2 has multiple intents, thus we treat them as multi-classification and multi-label classification problems respectively. 
We use the clean transcriptions of train sets to train intent recognition models, and use ASR results of test sets to evaluate model performance. We use our ASR system for FSC. Since DSTC-2 doesn't come with audio records, we use the self-contained ASR results recognized for testing. 

%CatSLu is a dataset similar to DSTC-2, it contains original audio recordings. Follow similar instructions, we So we can use our own ASR-engine to get asr transcription.

\subsubsection{Settings}

We use fasttext \cite{joulin2016bag} for intent classification. We use softmax loss for FSC and one-vs-all loss for DSTC-2. We set word vector dimension to 20 and wordNgram to 2. For DSTC-2, we set the probability threshold to 0.5 and choose up to 3 intents. We use accuracy to evaluate performance of FSC and  multi-label accuracy/F1-score \cite{sorower2010literature} for DSTC-2. We randomly run 5 times and report the average score. 

We use EDA \cite{wei2019eda} as another baseline to compare. EDA is a general data augmentation method for text classification. It adopts simple operations like insert/replace/drop/swap on clean sentences without considering ASR errors. By contrast, our method is designed for dealing with ASR errors specifically. We generate 4 noisy samples for each of the clean sentences.

\subsubsection{Reuslt}
\begin{table}[h]
	\centering
	\begin{tabular}{l|c|cc|cc}
		\toprule
		& FSC & \multicolumn{2}{c}{DSTC2-live*}  & \multicolumn{2}{c}{DSTC2-batch*}  \\
		\midrule
		& Acc & Acc          & F1       & Acc          & F1          \\
		\midrule
		clean    &     93.8 &     84.2  &  85.1 &    81.2         &     82.3         \\
		+noisy    &      98.5   &     86.5  &     87.5  &83.7     &    84.8   \\
		\midrule
		+EDA     &     94.8   &84.5&85.4&    81.6          &      82.7       \\
		+RS 	 &        94.1	&84.4&85.3&   81.4           &    82.4          \\
		+SS      &       94.6   &82.5&83.6&    80.1          &      81.3                     \\
		\midrule
		+Ours     &  \textbf{97.0}  &\textbf{85.2}&\textbf{86.0}&  \textbf{82.3}    &   \textbf{83.4}      \\
		\bottomrule
	\end{tabular}
	\caption{Accuracy (\%) and F1-score on ASR results of FSC and DSTC-2 test sets. * DSTC-2 contains two sets of ASR results recognized by the ASR system in live mode and a less accurate batch mode. We conduct experiment on both. +\textit{noisy} means adding the ASR results of train sets, which act as an upper-bound of model performance.}
	\label{table:slu_results}
\end{table}

The main results are shown in Table \ref{table:slu_results}. Our method significantly improves the performance on ASR test sets and outperforms other methods. 

EDA consistently improves performance on both FSC and DSTC-2. RS and SS only show slight improvements on FSC. They downgrade performance on DSTC-2. Our method improves performance on all test sets by a large margin. From the improvements on DSTC-2, which uses a different ASR system than ours, we can conclude the same as previous section (section 4.3), that our method can benefit model robustness to various ASR errors. Augmenting the training data with ASR results of audio recordings yields the best performance as ``+noisy'' shows. However, such data is expensive to obtain in production.

\section{Related Work}
It is necessary to improve the robustness of NLP models against ASR errors in speech-enabled systems. Prior work mainly focuses on these directions. 
\subsubsection{Data augmentation}
Data augmentation is simple and attractive because it does not modify the existing model architecture or introduce additional latency during inference time.

\citet{li2018improving} proposed four strategies to craft ASR-specific noise training examples mainly based on word substitution, among which homophone-based substitution achieves the best performance.

\citet{tsvetkov2014augmenting} proposed a rule-based method to firstly transform a word n-gram to its phonetic sequence, and then apply random edits on that phonetic sequence, and finally transform back to another word n-gram. 

\citet{wang2020data} used a data-driven approach rather than heuristic rules to simulate ASR errors. First, it collects pairs of ASR hypothesis and corresponding reference text. Then these pairs are aligned at word-level by minimizing the Levenshtein distance. Then it counts n-gram confusion frequencies from these aligned pairs. Then, the confusion matrix is used to replace n-grams in the clean text to generate noisy text. Moreover, to make the generated noise match the ASR's error distribution it uses a heuristic to control the substitution process.

Our work is mostly related to \cite{wang2020data}, however, we use a novel language generator to generate noisy samples instead of simple substitution. 
\subsubsection{Feature enhancement}
Some work uses phone features to enhance input representation. \citet{li2018improving,liu2018robust} adopted phone features by concating/adding Chinese Pinyin embedding to token embedding in the input layer of transformer model to make Zh-En translation model more robust to homophone errors.
\subsubsection{End-to-End approach}
The end-to-end approach seems promising to handle error propagation in traditional pipelined systems. The main issue is training data in the form of speech paired with downstream task ground truth, eg. translation target or intent/slot labels is extremely rare. Some works \cite{lugosch2019speech, 9053281, wang2020bridging} attempted to use end-to-end training for SLT/SLU task. \citet{wang2020covost,lugosch2019speech,di2019must} contributed some useful data resources to support end-to-end research. However, for now, pipelined systems are still mainstream in production.  

\subsubsection{ASR error correction and n-best reranking}
\citet{9053213} used an error correction module to correct errors in ASR outputs before sending them to downstream NLP models. \citet{li2020improving} tried to leverage n-best ASR hypotheses to deal with ASR noises. \citet{corona2017improving} reranked n-best ASR hypotheses using language models and sent the 1-best hypothesis to downstream NLP models.

\section{Conclusion}
Speech-enabled systems are fragile to ASR noise. In this paper, We propose a simple yet effective data augmentation
method to improve the robustness of NLP systems against ASR errors. Experimental results show that our method significantly improves the performance of machine translation tasks and intent classification tasks compared to previous data augmentation methods. In the future, we would like to explore more
powerful PLMs proposed in recent years (Lewis et al. 2019; Bao et al. 2020) for better noise generation, and evaluate our method on more languages such as Chinese and German.

\bibliographystyle{aaai21}
\bibliography{aaai2021}

\end{document}